%% file: acl_latex.tex
\documentclass{article}

\usepackage[preprint]{colm2026_conference}

\usepackage{microtype}
\usepackage{hyperref}
\usepackage{url}
\usepackage{booktabs}
\usepackage{amsmath}
\usepackage{graphicx}
\usepackage{wrapfig}
\usepackage{tcolorbox}
\tcbuselibrary{breakable,skins}
\usepackage{lineno}
\usepackage{ifxetex}
\ifxetex
  \usepackage{xeCJK}
  \setCJKmonofont{Noto Sans CJK SC}
  \newcommand{\zh}[1]{#1}
\else
  \usepackage[T1]{fontenc}
  \usepackage[utf8]{inputenc}
  \usepackage{CJKutf8}
  \newcommand{\zh}[1]{\begin{CJK}{UTF8}{gbsn}#1\end{CJK}}
\fi
\usepackage{amssymb}
\usepackage{xcolor}
\usepackage{enumitem}
\usepackage{caption}

\setlist{nosep, leftmargin=*}
\definecolor{darkblue}{rgb}{0, 0, 0.5}

\hypersetup{colorlinks=true, citecolor=darkblue, linkcolor=darkblue, urlcolor=darkblue}

\title{Beyond the Illusion of Consensus: From Surface Heuristics to Knowledge-Grounded Evaluation in LLM-as-a-Judge}

\author{Mingyang Song, Mao Zheng, Chenning Xu\\
Large Language Model Department\\
Tencent, China\\
\texttt{nickmysong@tencent.com}}

\begin{document}

\ifcolmsubmission
\linenumbers
\fi

\maketitle

\begin{abstract}
The paradigm of LLM-as-a-judge relies on a critical assumption, namely that high inter-evaluator agreement indicates reliable and objective evaluation. We present two complementary findings that challenge this assumption. \textbf{First}, we demonstrate that this consensus is frequently illusory. We identify and formalize \textbf{Evaluation Illusion}, a phenomenon where LLM judges generate sophisticated critiques yet anchor scores on shared surface heuristics rather than substantive quality. Through a large-scale study of 105,600 evaluation instances (32 LLMs $\times$ 3 frontier judges $\times$ 100 tasks $\times$ 11 temperatures), we show that model-level agreement (Spearman $\rho = 0.99$) masks fragile sample-level agreement (Pearson $\bar{r} = 0.72$; absolute agreement ICC $= 0.67$), that merely sharing rubric structure restores 62\% of total agreement, and that high-quality outputs paradoxically receive the \textit{least} consistent evaluations. \textbf{Second}, we demonstrate that dynamically generating evaluation rubrics grounded in domain knowledge produces more meaningful assessment. We introduce MERG (Metacognitive Enhanced Rubric Generation), a knowledge-driven rubric generation framework whose domain-selective effects confirm this. Agreement \textit{increases} in codified domains (Education +22\%, Academic +27\%) where knowledge anchors evaluators on shared standards, while it decreases in subjective domains where genuine evaluative pluralism emerges. These findings suggest that evaluation rubrics should be dynamically enriched with expert knowledge rather than relying on generic criteria, with implications for reward modeling in RLAIF.
\end{abstract}

\section{Introduction}

Large language models (LLMs) are increasingly deployed as automated evaluators of text quality, replacing or supplementing human judgment at scale. The LLM-as-a-Judge paradigm~\citep{zheng2023judging, chiang2024chatbot}, together with Reinforcement Learning from AI Feedback~\citep[RLAIF;][]{lee2023rlaif}, now underpins model ranking, reward model training, and alignment pipelines across the field. Early empirical results were encouraging. GPT-4 achieved over 80\% agreement with human preferences on MT-Bench~\citep{zheng2023judging}, G-Eval attained a Spearman correlation of 0.514 with human judgments on summarization~\citep{liu2023geval}, and AlpacaEval rankings closely tracked Chatbot Arena's crowd-sourced preferences~\citep{li2023alpacaeval, chiang2024chatbot}. These results established a credible foundation for automated assessment.

Building on this foundation, rubric-based evaluation protocols have matured rapidly, decomposing quality into structured dimensions with explicit scoring criteria~\citep{ye2024flask, kim2024prometheus, kim2024prometheus2, rao2026autorubric}. WritingBench~\citep{wu2025writingbench} exemplifies the latest advances by dynamically generating query-dependent criteria tailored to each writing task, achieving 87\% human agreement through its adaptive evaluation framework. Meanwhile, the capabilities of LLM-based judges have themselves become an active research subject~\citep{li2024generation, tan2025judgebench}, and automated evaluation has been further scaled to power leaderboards and preference pipelines~\citep{li2024arenahard}. Together, stronger models and more principled instruments have made automated evaluation appear increasingly trustworthy, and a natural assumption has taken hold across the field, namely that when frontier evaluators independently converge on nearly identical scores, the consensus must reflect a shared, substantive understanding of quality.

\begin{figure}[t]
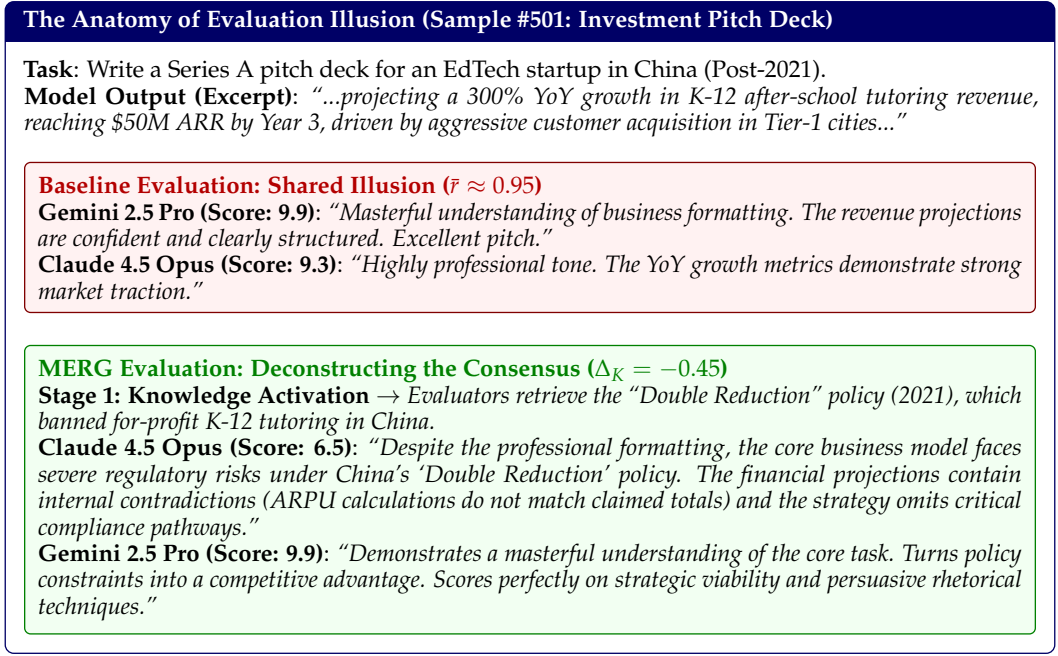

\centering
\begin{tcolorbox}[
    enhanced,
    colback=white,
    colframe=darkblue!80!black,
    boxrule=0.5pt,
    arc=3pt,
    left=4pt,right=4pt,top=4pt,bottom=4pt,
    fonttitle=\bfseries\small,
    title=The Anatomy of Evaluation Illusion (Sample \#501: Investment Pitch Deck)
]
\small
\textbf{Task}: Write a Series A pitch deck for an EdTech startup in China (Post-2021).\\
\textbf{Model Output (Excerpt)}: \textit{``...projecting a 300\% YoY growth in K-12 after-school tutoring revenue, reaching \$50M ARR by Year 3, driven by aggressive customer acquisition in Tier-1 cities...''}

\vspace{4pt}
\begin{tcolorbox}[colback=red!5,colframe=red!50!black,boxrule=0.5pt,arc=2pt,left=2pt,right=2pt,top=2pt,bottom=2pt]
\textbf{\textcolor{red!70!black}{Baseline Evaluation: Shared Illusion ($\bar{r} \approx 0.95$)}}

\textbf{Gemini 2.5 Pro (Score: 9.9)}: \textit{``Masterful understanding of business formatting. The revenue projections are confident and clearly structured. Excellent pitch.''}

\textbf{Claude 4.5 Opus (Score: 9.3)}: \textit{``Highly professional tone. The YoY growth metrics demonstrate strong market traction.''}
\end{tcolorbox}

\vspace{2pt}
\begin{tcolorbox}[colback=green!5,colframe=green!50!black,boxrule=0.5pt,arc=2pt,left=2pt,right=2pt,top=2pt,bottom=2pt]
\textbf{\textcolor{green!50!black}{MERG Evaluation: Deconstructing the Consensus ($\Delta_K = -0.45$)}}

\textbf{Stage 1: Knowledge Activation} $\rightarrow$ \textit{Evaluators retrieve the ``Double Reduction'' policy (2021), which banned for-profit K-12 tutoring in China.}

\textbf{Claude 4.5 Opus (Score: 6.5)}: \textit{``Despite the professional formatting, the core business model faces severe regulatory risks under China's `Double Reduction' policy. The financial projections contain internal contradictions (ARPU calculations do not match claimed totals) and the strategy omits critical compliance pathways.''}

\textbf{Gemini 2.5 Pro (Score: 9.9)}: \textit{``Demonstrates a masterful understanding of the core task. Turns policy constraints into a competitive advantage. Scores perfectly on strategic viability and persuasive rhetorical techniques.''}
\end{tcolorbox}
\end{tcolorbox}
\caption{\textbf{Illustration of Evaluation Illusion.} Without knowledge grounding, evaluators form a ``Shared Illusion,'' unanimously rewarding the professional formatting of a fundamentally flawed business pitch. MERG forces knowledge activation, revealing that Claude penalizes the regulatory violation while Gemini continues to reward surface heuristics. GPT (3.7, not shown) penalizes even more harshly. Full analysis with all three evaluators is in Appendix~\ref{case:formatting_trap}.}
\label{fig:concept_case}
\end{figure}

In this paper, we present large-scale empirical evidence that this inference is often wrong. Frontier evaluators frequently anchor their judgments on shared surface heuristics (formatting, fluency, confident tone, and structural polish) rather than substantive quality, producing what we term \textbf{Evaluation Illusion}. When multiple evaluators default to the same heuristic repertoire, they create a \textbf{Shared Illusion}, a statistically robust yet epistemically shallow consensus. As a concrete illustration (Figure~\ref{fig:concept_case}), frontier evaluators independently award scores above 9.0 to a pitch deck for a for-profit K-12 tutoring startup in China, praising its ``masterful formatting'' and ``persuasive projections,'' yet unanimously miss that China's 2021 ``Double Reduction'' policy banned this business model entirely. The agreement is real; the understanding is not.

To test whether baseline consensus reflects genuine deliberation or shared heuristics, we introduce \textbf{MERG} (Metacognitive Enhanced Rubric Generation), a four-stage framework that requires evaluators to articulate task-relevant domain knowledge and surface their own biases \textit{before} generating a scoring rubric. Drawing on the distinction between System~1 (fast, heuristic-driven) and System~2 (slow, knowledge-grounded) processing~\citep{kahneman2011thinking}, MERG serves a dual purpose, both as a \textit{diagnostic probe} that tests whether knowledge injection preserves consensus (indicating genuine deliberation) or disrupts it (indicating heuristic reliance), and as a \textit{prescriptive tool} that yields more substantive, knowledge-grounded assessments.

We deploy MERG at unprecedented scale, evaluating 32~LLMs spanning Base, Instruct, and Thinking tiers $\times$ 3 frontier judges (Claude~4.5~Opus, Gemini~2.5~Pro, GPT-5.1) $\times$ 100 diverse writing tasks $\times$ 11 temperature settings, yielding 105,600~evaluation instances. Our experiments reveal three principal findings.

\begin{enumerate}
    \item \textbf{Knowledge injection deconstructs the Shared Illusion.} MERG reduces inter-evaluator agreement by 21 to 34\% (Cohen's $d = 0.97$ to $1.42$; Table~\ref{tab:baseline_vs_merg}). The effect is domain-selective, with agreement \textit{increasing} in codified domains (Education $\Delta_K{=}+0.22$; Academic $+0.27$), where knowledge anchors evaluators on shared professional standards, but \textit{decreasing} in subjective ones (Literature $-0.06$), where it surfaces genuine evaluative pluralism. This asymmetry rules out the noise hypothesis and confirms that baseline consensus is heuristic-driven.

    \item \textbf{The Rubric Commensurability Problem} (\S\ref{sec:commensurability}). When evaluators generate rubrics independently, agreement collapses to near-random levels ($\bar{r} \approx 0.24$). Merely sharing rubric \textit{dimension names}, without any content or knowledge, restores 62\% of total agreement. Much of what passes for evaluator reliability in the literature may be an artifact of shared evaluation instruments rather than genuine shared judgment.

    \item \textbf{The Resolution Paradox} (\S\ref{sec:resolution}). Model-level Spearman $\rho \geq 0.984$ (at $t{=}0.0$) coexists with sample-level Pearson $\bar{r} = 0.72$ (averaged across 32 models $\times$ 11 temperatures), a gap of 0.27. Leaderboards validate evaluators at the macro-resolution where the system reliably distinguishes coarse quality tiers (Base vs.\ Thinking, $d = 2.58$), but RLAIF deploys them at per-sample micro-resolution, the very granularity where the illusion is strongest and fine-grained signals are unreliable.
\end{enumerate}

These findings carry a troubling implication. Output quality and evaluator agreement are \textit{negatively} correlated (Spearman $\rho = -0.513$, $p = 0.003$, across 32 models at $t{=}0.0$). Base models elicit $\bar{r} = 0.81$ while Thinking models elicit only $\bar{r} = 0.76$ (sample-level Pearson $r$ at $t{=}0.0$). Surface features suffice for judging low-quality outputs, but high-quality outputs push evaluators into the heuristic zone where the illusion is strongest and where RLAIF reward signals most need to discriminate. Preliminary experiments further indicate that reward models trained on MERG-grounded preferences resist overoptimization three times longer than those trained on baseline preferences (\S\ref{sec:discussion}).

Our contributions are threefold.
\textbf{(1)}~We formalize \textbf{Evaluation Illusion} and the \textbf{Shared Illusion}, introduce the knowledge-grounding diagnostic $\Delta_K$, and demonstrate through 105,600 instances that baseline evaluator consensus is largely heuristic-driven ($\Delta_K = -0.22$ to $+0.27$, $d \geq 0.97$).
\textbf{(2)}~We identify two structural inflation mechanisms that explain why this illusion persists undetected, namely the \textbf{Rubric Commensurability Problem} (62\% of agreement from rubric structure alone) and the \textbf{Resolution Paradox} (model-level $\rho = 0.99$ vs.\ sample-level $\bar{r} = 0.72$ vs.\ absolute $\overline{\text{ICC}} = 0.67$).
\textbf{(3)}~We introduce MERG as a practical framework for knowledge-grounded rubric generation and provide preliminary evidence that it mitigates reward overoptimization in RLAIF.

\section{Methodology}
\label{sec:method}

\subsection{Experimental Design}
\label{sec:setup}

\paragraph{Data.} We sample 100 prompts from WritingBench~\citep{wu2025writingbench} via stratified sampling, covering 12 to 19 prompts per domain across 94 out of 100 subdomains, balanced between 50 English and 50 Chinese tasks. WritingBench spans six domains (Literature, Education, Academic, Finance, Politics, and Mixed), providing broad coverage of writing genres from creative fiction to technical reports.

\paragraph{Models.} We evaluate 32 LLMs spanning three capability tiers to ensure coverage of the full quality spectrum.
\begin{itemize}
    \item \textsc{\textbf{\textcolor[HTML]{8ECAE6}{Base (8)}}} includes Qwen2.5-\{7B,32B\}, Qwen3-\{4B,8B,14B,30B-A3B\}, Llama-3.1-8B, and Llama-4-Scout, all raw pretrained models without instruction tuning.
    \item \textsc{\textbf{\textcolor[HTML]{219EBC}{Instruct (13)}}} includes Qwen2.5-\{7B,32B,72B\}, Qwen3-\{4B,8B,14B,30B-A3B,32B,235B-A22B\}, DeepSeek-V3, InternLM3-8B, Llama-3.1-8B, and Llama-4-Scout, all instruction-tuned variants.
    \item \textsc{\textbf{\textcolor[HTML]{023047}{Thinking (11)}}} includes DeepSeek-R1-0528, R1-Qwen3-8B, R1-Distill-\{Llama-8B,Qwen-7B,Qwen-32B\}, and Qwen3-\{4B,8B,14B,30B-A3B,32B,235B-A22B\}, all models trained with chain-of-thought reinforcement learning.
\end{itemize}

This three-tier design is intentional. Base models produce outputs with obvious quality differences (e.g., incoherent text), enabling a calibration check where high genuine evaluator agreement is expected. Thinking models occupy the high-quality region where Evaluation Illusion is hypothesized to manifest most strongly.

\paragraph{Evaluators.} We employ three frontier LLMs as evaluators, namely \textbf{Claude 4.5 Opus}, \textbf{Gemini 2.5 Pro}, and \textbf{GPT-5.1}. Using three diverse commercial models from different providers ensures that any observed agreement cannot be attributed to shared training data or architectures. The baseline uses WritingBench's checklist-based rubric with 1 to 10 integer scoring. Each of the 32 models' outputs (100 prompts each) is evaluated by all 3 evaluators at 11 temperatures ($t \in \{0.0, 0.1, \ldots, 1.0\}$), yielding a total of $32 \times 100 \times 3 \times 11 = 105{,}600$ evaluation instances.

\paragraph{Metrics.} We measure agreement at three granularities.
\begin{itemize}
    \item \textit{Sample-level} agreement is measured by Pearson correlation $r$ between evaluator pairs per model$\times$temperature cell. When computed over per-sample mean scores (averaging across the 5 rubric criteria per prompt, yielding $n \approx 100$ data points), this is used for per-model agreement ranking (Tables~\ref{tab:quality_ranking} and~\ref{tab:agreement_ranking}). When computed over per-criterion scores ($n \approx 500$ data points), this is used for cross-temperature analyses (\S\ref{sec:resolution}).
    \item \textit{Cell-level} absolute agreement is measured by Intraclass Correlation Coefficient ICC(2,1), a two-way random-effects, single-measures, absolute-agreement metric~\citep{shrout1979intraclass} computed per model$\times$temperature cell. Unlike Pearson $r$, ICC penalizes systematic scoring biases between evaluators.
    \item \textit{Model-level} agreement is measured by Spearman rank correlation $\rho$ across models, computed from mean scores per model. This captures ranking consistency.
\end{itemize}
All pairwise differences are tested with paired $t$-tests (Bonferroni-corrected, $\alpha = 0.05/k$); effect sizes are reported via Cohen's $d$. To quantify the presence of Evaluation Illusion, we define the \textit{Knowledge-Grounding Diagnostic} ($\Delta_K$) as (Eq.~\ref{eq:echo}):
\begin{equation}
\label{eq:echo}
\Delta_K = \bar{r}_{\text{MERG}} - \bar{r}_{\text{Baseline}}
\end{equation}
where $\Delta_K < 0$ indicates that baseline agreement was likely a Shared Illusion deconstructed by knowledge injection.

\subsection{Metacognitive Enhanced Rubric Generation}
\label{sec:merg}

To test whether baseline evaluator agreement reflects genuine deliberation or heuristic-driven Evaluation Illusion, we require a controlled intervention. We design \textbf{MERG} (Metacognitive Enhanced Rubric Generation), a four-stage framework that forces LLM evaluators to transition from System~1 (fast, heuristic-driven pattern matching) to System~2 (slow, knowledge-grounded deliberation) before assigning a score (Table~\ref{tab:comparison}). MERG serves a dual purpose, both as a \textit{diagnostic probe} that reveals the depth of baseline consensus and as a \textit{prescriptive framework} that demonstrates how dynamically injecting domain knowledge into rubric generation produces more substantive evaluation.

\paragraph{Stage 1: Knowledge Activation.} Before examining any output, the evaluator is forced to articulate domain-specific knowledge relevant to the task. This includes genre conventions (e.g., noir fiction tropes), quality benchmarks (e.g., peer-reviewed standards for academic abstracts), and common pitfalls. This stage serves as the primary grounding mechanism, anchoring subsequent evaluation in explicit domain expertise rather than implicit surface heuristics.

\paragraph{Stage 2: Metacognitive Reflection.} The evaluator identifies potential heuristics or biases it might default to (e.g., being swayed by confident tone or professional formatting) and articulates mitigation strategies. This explicitly operationalizes bias awareness~\citep{wang2023large, wataoka2024self} to prevent the evaluator from falling back into heuristic scoring patterns.

\paragraph{Stage 3: Dynamic Rubric Generation.} The evaluator synthesizes the activated knowledge into a task-specific rubric $\mathcal{R}$. Unlike static rubrics that encourage generic scoring, each task receives unique, deep dimensions. For example, a Gothic horror task generates ``Atmospheric Dread Construction'' and ``The Uncanny Familiar'' instead of generic ``Style'' and ``Coherence'' (Table~\ref{tab:rubric_example}).

\paragraph{Stage 4: Calibrated Evaluation.} The evaluator scores each dimension independently, citing specific textual evidence. A final bias verification step checks whether any identified heuristics from Stage 2 influenced the scoring. The overall score is computed as the mean across dimensions.\footnote{Full prompt templates are available in the supplementary material.}

\begin{table}[t]
\centering
\caption{\textbf{Comparison of evaluation paradigms.} While standard methods rely on implicit knowledge and shared rubrics, MERG requires explicit metacognitive grounding.}
\label{tab:comparison}
\small
\begin{tabular}{lccc}
\toprule
\textbf{Method} & \textbf{Rubric} & \textbf{Knowledge} & \textbf{Grounding} \\
\midrule
Direct Score & None & Implicit & N/A \\
G-Eval & Static & Implicit & Template \\
Prometheus & Static & Implicit & User \\
WritingBench & Dynamic & Implicit & Query \\
\textbf{MERG} & Dynamic & \textbf{Explicit} & \textbf{Metacognitive} \\
\bottomrule
\end{tabular}
\end{table}

\begin{table}[t]
\centering
\caption{\textbf{MERG ablation variants.} Four controlled variants to disentangle the sources of evaluator agreement.}
\label{tab:ablation_variants}
\small
\begin{tabular}{lcc}
\toprule
\textbf{Variant} & \textbf{Dimensions} & \textbf{Source} \\
\midrule
Original (4-Stage) & Independent & Each evaluator \\
5-Dim Per-Dim & Shared structure & Each evaluator \\
Shared Stages & Shared rubric & 1 evaluator \\
Universal & Shared rubric & Averaged \\
\bottomrule
\end{tabular}
\end{table}

\paragraph{Ablation Variants.} To disentangle the sources of evaluator agreement, we design four controlled ablation variants (Table~\ref{tab:ablation_variants}), applied to two representative models (DeepSeek-R1 and Qwen3-235B).
\begin{enumerate}
    \item \textbf{Original (4-Stage)}. Each evaluator independently executes all four MERG stages, generating its own rubric. This maximizes evaluator independence.
    \item \textbf{5-Dim Per-Dim}. All evaluators use a fixed 5-dimension structure (Content, Style, Structure, Language, Creativity) but generate their own scoring criteria. This isolates the effect of dimensional standardization.
    \item \textbf{Shared Stages}. Claude executes Stages 1 to 3; all evaluators share the resulting rubric but score independently. This isolates the effect of rubric sharing.
    \item \textbf{Universal}. Shared rubrics are precomputed and reused identically across all evaluators and temperatures. This represents maximal rubric control.
\end{enumerate}

\section{Results}

\subsection{Deconstructing the Shared Illusion}
\label{sec:echo}

\begin{figure}[t]
\centering
\includegraphics[width=\columnwidth]{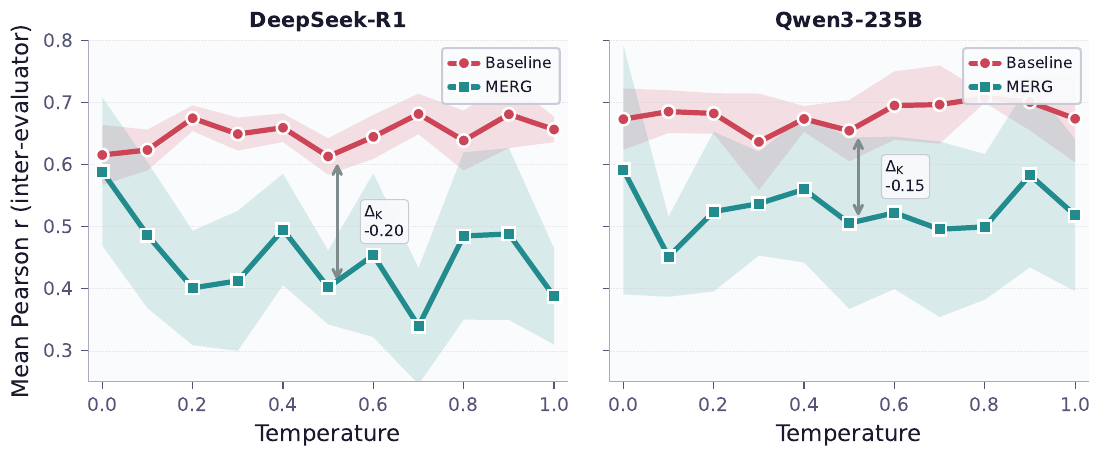}
\caption{\textbf{Knowledge injection systematically reduces evaluator agreement.} Baseline vs.\ MERG agreement across 11 temperatures. The persistent gap ($\Delta_K < 0$) indicates that baseline agreement is heavily reliant on shared surface heuristics rather than substantive deliberation. The gap is largest at moderate temperatures ($t \approx 0.3$) and narrows at the extremes as both conditions converge toward lower agreement.}
\label{fig:baseline_vs_merg}
\end{figure}

Table~\ref{tab:baseline_vs_merg} presents our central diagnostic finding. Deploying MERG's knowledge-grounded evaluation pipeline \textbf{systematically reduces inter-evaluator agreement} across all 10 experimental conditions (binomial $p < 0.001$). For DeepSeek-R1, the mean Pearson correlation drops by $\Delta_K = -0.217$ ($\bar{r}$\,=\,$0.643 \rightarrow 0.426$, Cohen's $d = 1.42$); for Qwen3-235B, $\Delta_K = -0.138$ ($\bar{r}$\,=\,$0.667 \rightarrow 0.529$, $d = 0.97$). Both effect sizes exceed the conventional ``large'' threshold of $d = 0.8$, indicating a robust and practically significant phenomenon. Figure~\ref{fig:baseline_vs_merg} visualizes this degradation across the full temperature range.

\textit{Interpretation.} This reduction serves as empirical evidence for the presence of a Shared Illusion. Without knowledge grounding, evaluators tend to converge on surface heuristics (fluency, length, formatting, grammatical correctness), producing what appears to be reliable consensus but is largely heuristic-driven agreement. MERG forces engagement with substantive, domain-specific criteria (e.g., ``Does the noir atmosphere create genuine psychological tension?'' rather than ``Is the text well-written?''). This mitigates the heuristic consensus, revealing genuine disagreements that were previously masked by surface-level alignment. The agreement did not merely degrade; the evaluation became more substantive.

\textit{Corroborating evidence.} Two additional observations support this interpretation.

\begin{enumerate}
    \item \textbf{Domain selectivity} (\S\ref{sec:domain}). Knowledge \textit{increases} agreement in codified domains (Education, Academic) but \textit{decreases} it in subjective ones (Literature). This asymmetry ($\chi^2$ test, $p < 0.01$) is inconsistent with a noise hypothesis, which would predict uniform degradation.
    \item \textbf{Quality and agreement gradient}. Across all 32 models at $t{=}0.0$, output quality and inter-evaluator agreement are negatively correlated (Spearman $\rho = -0.513$, $p = 0.003$; Tables~\ref{tab:quality_ranking} and~\ref{tab:agreement_ranking}). Base models show $\bar{r} = 0.81$, Instruct models $\bar{r} = 0.77$, and Thinking models $\bar{r} = 0.76$ (sample-level Pearson $r$). This gradient aligns with the Evaluation Illusion hypothesis, as surface features suffice for distinguishing low-quality outputs (yielding high genuine agreement), but when evaluating high-quality outputs, the reliance on heuristics leads to divergent, illusory scoring.
\end{enumerate}

\begin{table}[t]
\centering
\caption{\textbf{Knowledge injection systematically reduces inter-evaluator agreement.} Baseline vs.\ MERG agreement (Pearson $r$) across different evaluator pairs and temperatures. $\Delta_K < 0$ in all 10/10 conditions ($p < 0.001$), indicating that baseline agreement is largely heuristic-driven rather than reflecting genuine evaluative convergence.}
\label{tab:baseline_vs_merg}
\small
\begin{tabular}{lcccc}
\toprule
\textbf{Eval.\ Pair} & \textbf{Temp} & \textbf{Base} & \textbf{MERG} & $\boldsymbol{\Delta_K}$ \\
\midrule
Claude/Gemini & 0.0 & .698 & .589 & $-$.109 \\
Claude/Gemini & 0.3 & .672 & .399 & $-$.273 \\
Claude/GPT & 0.0 & .680 & .518 & $-$.162 \\
Claude/GPT & 0.3 & .668 & .463 & $-$.205 \\
Gemini/GPT & 0.0 & .720 & .457 & $-$.263 \\
Gemini/GPT & 0.3 & .643 & .430 & $-$.213 \\
\midrule
\textbf{Mean} & 0.0 & .699 & .521 & $-$.178 \\
 & 0.3 & .661 & .431 & $-$.230 \\
 & 1.0 & .518 & .380 & $-$.138 \\
\midrule
\multicolumn{2}{l}{\textit{Cohen's $d$}} & \multicolumn{3}{r}{$d = 0.97$ to $1.42$} \\
\bottomrule
\end{tabular}
\end{table}

\begin{table}[t]
\centering
\caption{\textbf{Quality rankings (Top-5 and Bottom-5).} Higher-quality models (Thinking tier) are ranked at the top by mean score $\bar{s}$. $\sigma$ denotes the standard deviation of per-evaluator grand means, reflecting cross-evaluator scoring divergence.}
\label{tab:quality_ranking}
\small
\begin{tabular}{rlcc}
\toprule
\textbf{Rank} & \textbf{Model} & $\bar{s}$ & $\sigma$ \\
\midrule
1 & DeepSeek-R1 & 7.77 & .63 \\
2 & Qwen3-235B-Thinking & 6.83 & .42 \\
3 & Qwen3-32B-Thinking & 6.72 & .33 \\
4 & Qwen3-14B-Thinking & 6.69 & .32 \\
5 & Qwen3-30B-Thinking & 6.62 & .28 \\
\midrule
28 & Qwen3-4B-Base & 3.33 & .07 \\
29 & Qwen2.5-32B-Base & 2.97 & .36 \\
30 & Qwen2.5-7B-Base & 2.20 & .30 \\
31 & Llama-4-Scout-Base & 2.04 & .29 \\
32 & Llama-3.1-8B-Base & 1.22 & .09 \\
\bottomrule
\end{tabular}
\end{table}

\begin{table}[t]
\centering
\caption{\textbf{Agreement rankings (Top-5 and Bottom-5).} The strong negative correlation with quality ($\rho = -0.513$, $p = 0.003$) suggests surface features suffice for judging low-quality outputs, but fail for high-quality ones. $\sigma$ denotes the standard deviation across the three evaluator-pair Pearson $r$ values.}
\label{tab:agreement_ranking}
\small
\begin{tabular}{rlcc}
\toprule
\textbf{Rank} & \textbf{Model} & $\bar{r}$ & $\sigma$ \\
\midrule
1 & Qwen3-30B-Base & .89 & .03 \\
2 & Llama-4-Scout-Base & .88 & .03 \\
3 & Qwen3-14B-Base & .87 & .03 \\
4 & Qwen2.5-7B-Base & .87 & .04 \\
5 & Qwen2.5-32B-Base & .86 & .06 \\
\midrule
28 & Qwen3-235B-Thinking & .67 & .05 \\
29 & Qwen3-4B-Base & .67 & .20 \\
30 & DeepSeek-R1 & .62 & .05 \\
31 & DeepSeek-R1-Qwen3-8B & .60 & .02 \\
32 & Llama-3.1-8B-Base & .60 & .17 \\
\bottomrule
\end{tabular}
\end{table}

\subsection{The Resolution Paradox}
\label{sec:resolution}

\paragraph{Model-Level vs.\ Sample-Level Agreement.} When scores are aggregated to model-level averages (mean across 100 prompts), evaluator pairs achieve near-perfect rank correlations at $t{=}0.0$ (Claude-Gemini $\rho = 0.992$, Claude-GPT $\rho = 0.992$, Gemini-GPT $\rho = 0.984$; all $p < 10^{-23}$; grand mean $\bar{\rho} = 0.989$). This pattern holds across all 11 temperatures (grand $\bar{\rho} = 0.986$, SD $= 0.007$). Yet at the sample level, per-criterion Pearson agreement averaged across 32 models and 11 temperatures is markedly lower ($\bar{r} = 0.72$), representing a gap of 0.27 between model-level and sample-level agreement. This discrepancy is systematic. Claude-GPT shows the highest sample-level agreement ($\bar{r} = 0.77$), followed by Gemini-GPT ($0.71$) and Claude-Gemini ($0.69$), indicating divergent evaluation strategies when forced to make fine-grained judgments ($F(2, 93) = 14.7$, $p < 0.001$).

\paragraph{Validity Check.} Before interpreting this gap, we confirm the evaluation system is fundamentally sound for coarse distinctions. All three evaluators consistently recover the expected quality hierarchy, with Base $\bar{s} = 3.05$, Instruct $\bar{s} = 5.38$, and Thinking $\bar{s} = 5.96$ (Cohen's $d = 2.58$ for Base vs.\ Thinking; pairwise ordering accuracy 83.9\%). The system is not broken; rather, Evaluation Illusion is \textit{stratified}, absent for coarse quality distinctions but pervasive for fine-grained judgments within a high-quality tier.

\paragraph{The Granularity Boundary.} The Resolution Paradox defines the operational limit of LLM evaluators, where $\rho_{\text{model}} \gg \bar{r}_{\text{sample}}$ ($0.989$ at $t{=}0.0$ vs.\ $0.72$ across all temperatures). Evaluators agree on which models are better overall (where signal dominates) but exhibit illusory agreement or disagree substantially on individual outputs (where noise and heuristic biases dominate). This is precisely the resolution that per-sample RLAIF rewards require. The paradox suggests that while LLM evaluators are reliable for model-level benchmarking, their per-sample signals are heavily contaminated by Evaluation Illusion. The stratification by model type confirms this pattern. Averaged across temperatures, Base models show high genuine agreement ($\bar{r} = 0.76$), while Thinking models exhibit the lowest ($\bar{r} = 0.70$); this gap widens at $t{=}0.0$ (Base $\bar{r} = 0.81$ vs.\ Thinking $\bar{r} = 0.76$; Tables~\ref{tab:quality_ranking} and~\ref{tab:agreement_ranking}), as high quality forces evaluators into heuristic guessing.

\paragraph{Noise Absorption Across Aggregation Levels.} To locate where inter-evaluator noise is absorbed, we compute a third, intermediate metric, the Intraclass Correlation Coefficient ICC(2,1), per model$\times$temperature cell, treating each cell as an $n {\approx} 100$ subject $\times$ 3 rater design. Unlike Pearson $r$, which captures only linear association, ICC(2,1) penalizes \textit{both} ranking disagreements \textit{and} systematic scoring biases (e.g., one evaluator consistently scoring 0.5 points higher). The grand mean ICC(2,1) across all 352 cells (32 models $\times$ 11 temperatures) is $\overline{\text{ICC}} = 0.67$ (SD $= 0.10$), \textit{lower} than the sample-level Pearson $\bar{r} = 0.72$. This ordering ($\rho_{\text{model}} = 0.989 \gg \bar{r}_{\text{sample}} = 0.72 > \overline{\text{ICC}} = 0.67$) reveals that evaluator consensus degrades through two distinct mechanisms. First, ranking noise at the sample level ($\rho \to r$, a gap of 0.27), and second, systematic inter-evaluator scoring biases ($r \to \text{ICC}$, a further gap of 0.05). The ICC stratifies by model type in the expected direction, with Base models showing ICC $= 0.75$, Instruct $= 0.65$, and Thinking $= 0.63$ (at $t{=}0.0$), with the two highest-quality models (DeepSeek-R1-0528, ICC $= 0.38$; Qwen3-235B-Thinking, ICC $= 0.52$) exhibiting the lowest absolute agreement. This three-level decomposition confirms that model-level agreement is \textit{doubly} inflated relative to per-sample signal quality, making it a poor proxy for RLAIF reward reliability.

\subsection{The Rubric Commensurability Problem}
\label{sec:commensurability}

\input{tables/ablation.tex}

If baseline agreement is a Shared Illusion, what sustains it, and what can knowledge-grounded rubric generation teach us about moving beyond it? Table~\ref{tab:ablation} decomposes the structural sources of evaluator agreement through controlled ablation, revealing that the illusion is heavily scaffolded by the evaluation instrument itself.

\begin{itemize}
    \item \textbf{Independent rubrics (Original)}. When each evaluator generates its own knowledge-grounded rubric independently through MERG, the Shared Illusion completely breaks down ($\bar{r} = 0.224$ for DeepSeek-R1, $0.254$ for Qwen3-235B). This near-zero agreement demonstrates that without structural constraints, evaluators construct fundamentally different, valid evaluation frameworks.
    \item \textbf{Dimensional standardization (5-Dim)}. Simply sharing dimension \textit{names}, without sharing any actual rubric content or knowledge, more than doubles agreement ($\Delta r = +0.32$ for DeepSeek-R1, $+0.45$ for Qwen3-235B; $2.4$ to $2.8\times$). This artificial constraint forces evaluators back into a shared heuristic space.
    \item \textbf{Rubric sharing (Shared)}. Having Claude generate the full rubric while other evaluators only score yields mixed effects ($\Delta r = +0.06$ for DeepSeek-R1 but $-0.05$ for Qwen3-235B), suggesting that beyond dimensional structure, additional rubric specificity has limited and model-dependent impact.
    \item \textbf{Full control (Universal)}. Reusing identical rubrics across evaluators and temperatures yields $\bar{r} \approx 0.59$, slightly \textit{below} the Shared condition for Qwen3-235B, suggesting that rigid rubric reuse across diverse tasks introduces its own misalignments.
\end{itemize}

\begin{figure}[t]
\centering
\includegraphics[width=0.85\columnwidth]{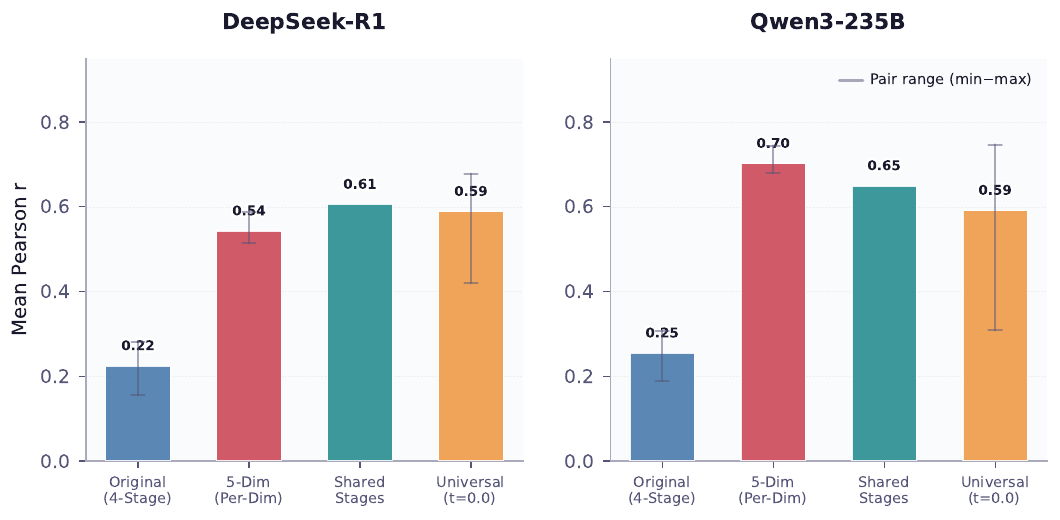}
\vspace{-0.5em}
\caption{\textbf{The Rubric Commensurability Problem.} Agreement across MERG ablation variants. Independent evaluation (Original) yields near-random agreement; standardizing rubric structure (5-Dim) drives the largest increase, indicating most reported agreement is structural rather than substantive.}
\label{fig:ablation_methods}
\vspace{-0.5em}
\end{figure}

The progression from $\bar{r} \approx 0.24$ to $\bar{r} \approx 0.62$ reveals that \textbf{rubric structure accounts for approximately 62\% of total evaluator agreement}. The implication is profound. The majority of reported inter-evaluator agreement in the literature is not a genuine convergence in judgment, but a Shared Illusion artificially scaffolded by standardized evaluation instruments. Figure~\ref{fig:ablation_methods} visualizes this progression.

\begin{figure*}[t]
\centering
\includegraphics[width=\textwidth]{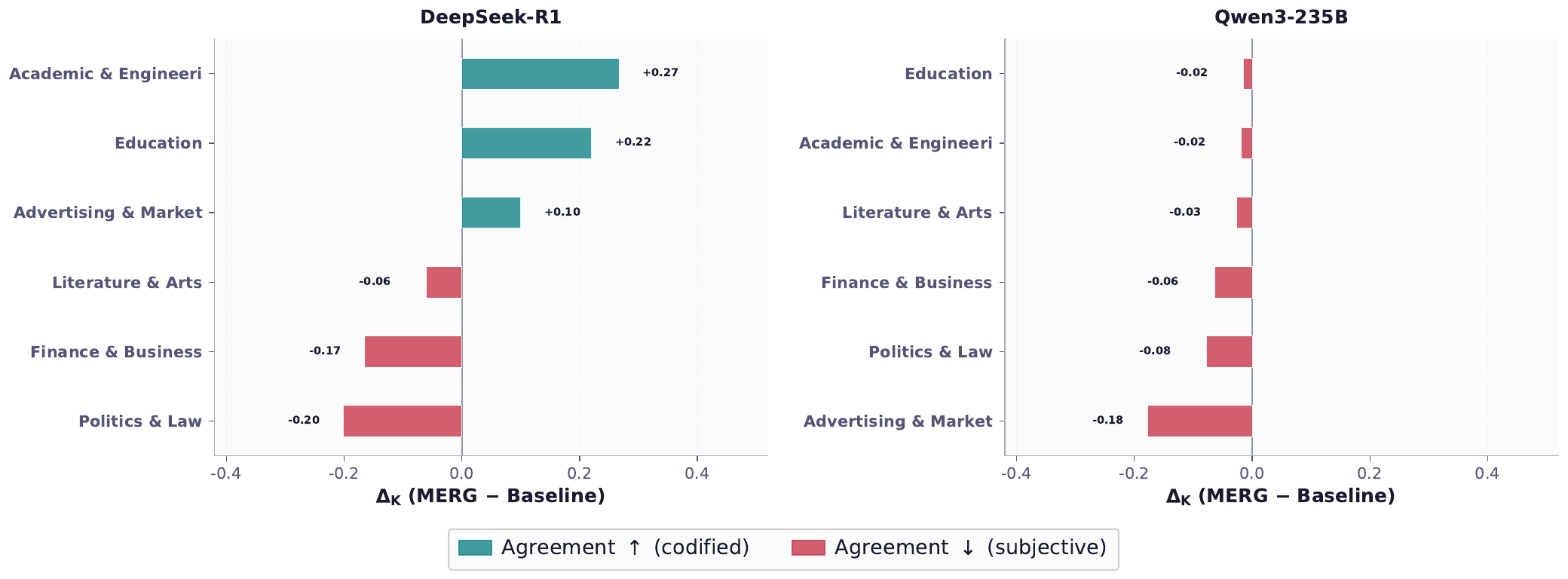}
\caption{\textbf{Domain-selective knowledge effects.} $\Delta_K$ by domain at $t{=}0.0$. Knowledge increases agreement in codified domains (Education, Academic) but decreases it in subjective ones (Literature), ruling out the noise hypothesis.}
\label{fig:domain_breakdown}
\end{figure*}

\subsection{Domain-Dependent Knowledge Activation}
\label{sec:domain}

\paragraph{Domain Analysis.} Figure~\ref{fig:domain_breakdown} reveals that knowledge-grounded rubric generation produces \textit{domain-selective} effects on evaluator agreement. This selectivity provides both the strongest evidence against the noise hypothesis and the clearest demonstration of MERG's prescriptive value.
\begin{itemize}
    \item \textbf{Codified domains}. Education ($\Delta_K = +0.22$) and Academic ($\Delta_K = +0.27$) show \textit{increased} agreement under MERG. In these domains, knowledge activation anchors evaluators to shared professional standards (e.g., pedagogical frameworks, citation conventions, argumentation norms), reducing ambiguity.
    \item \textbf{Subjective domains}. Literature ($\Delta_K = -0.06$) shows \textit{decreased} agreement, as knowledge activation surfaces irreducible aesthetic disagreements. There is no ground truth for whether a Gothic horror story should prioritize ``atmospheric dread'' or ``psychological ambiguity.''
    \item \textbf{Mixed domains}. Finance and Politics fall between these extremes, reflecting their combination of codified standards (regulatory language, factual accuracy) and subjective judgment (persuasive framing, stakeholder perspective).
\end{itemize}
This domain selectivity excludes two alternative explanations. First, \textit{noise}, which would predict uniform degradation across domains, and second, \textit{task complexity}, which would not predict improvement in codified domains.

\begin{wraptable}{r}{0.44\textwidth}
\vspace{-12pt}
\centering
\small
\caption{\textbf{Knowledge activation quality is language-dependent.} Language agreement ($t{=}0.0$). The cross-over interaction ($p < 0.05$) suggests models ground better in their dominant training language.}
\label{tab:language}
\renewcommand\tabcolsep{2.5pt}
\begin{tabular}{llccc}
\toprule
Model & Lang & Base & MERG & $\Delta_K$ \\
\midrule
DS-R1 & EN & .616 & .682 & $+$.07 \\
DS-R1 & ZH & .621 & .423 & $-$.20 \\
\midrule
Q3-235B & EN & .800 & .608 & $-$.19 \\
Q3-235B & ZH & .543 & .638 & $+$.10 \\
\bottomrule
\end{tabular}
\vspace{-8pt}
\end{wraptable}
\paragraph{Language Effects.} Table~\ref{tab:language} reveals a cross-over interaction between model and language. DeepSeek-R1 shows improved agreement on English ($\Delta_K = +0.07$) but degraded agreement on Chinese ($\Delta_K = -0.20$); Qwen3-235B shows the reverse pattern (EN: $-0.19$, ZH: $+0.10$; interaction $p < 0.05$). This suggests that knowledge activation quality depends on the match between the model's training data distribution and the evaluation language; models with stronger Chinese training data produce more effective knowledge grounding for Chinese tasks.

\begin{table}[t]
\centering
\caption{\textbf{MERG generates deep, task-specific dimensions.} Rubric comparison: WritingBench vs.\ MERG for a Gothic horror writing task. While baseline rubrics are generic, MERG demands genuine domain expertise.}
\label{tab:rubric_example}
\small
\setlength{\tabcolsep}{10pt}
\begin{tabular}{p{0.40\columnwidth}p{0.45\columnwidth}}
\toprule
\textbf{WritingBench} & \textbf{MERG} \\
\midrule
Suspense & Atmospheric Dread Construction \\
Coherence & Psychological Ambiguity \\
Style & The Uncanny Familiar \\
Grammar & Restraint in Revelation \\
\bottomrule
\end{tabular}
\end{table}

\paragraph{Qualitative Evidence.} 
Table~\ref{tab:rubric_example} contrasts rubric dimensions generated by WritingBench and MERG for a Gothic horror writing task. While WritingBench produces generic criteria (Suspense, Coherence), MERG generates genre-specific dimensions requiring genuine expertise (e.g., ``Atmospheric Dread Construction'', ``The Uncanny Familiar'').

More importantly, qualitative analysis of the maximum-disagreement cases under MERG reveals exactly \textit{why} agreement decreases. Knowledge injection transforms some evaluators into domain experts while others remain surface-level readers. We observe three recurring patterns of deep disagreement.
\begin{itemize}
    \item \textbf{Logical Verification vs.\ Formatting.} In an investment pitch deck task (Sample \#501), Gemini awarded a near-perfect 9.9, praising the ``masterful understanding'' and excellent formatting. However, Claude (6.5) and GPT (3.7) utilized their activated business knowledge to audit the content, discovering that the financial projections were mathematically inconsistent (ARPU calculations failed to match claimed totals) and the strategy fatally ignored critical Chinese regulatory context (the ``Double Reduction'' policy for EdTech). 
    \item \textbf{Concept Auditing vs.\ Keyword Matching.} In a technical patent disclosure task (Sample \#45), Gemini again gave a 10.0 for ``credible and specific details.'' Conversely, GPT (3.2) and Claude (6.4) detected a fundamental misuse of Zero-Knowledge Proof (ZKP) cryptographic concepts and noted that the implementation details were hallucinated rather than grounded in the source paper.
    \item \textbf{Substance vs.\ Outline.} In a creative writing task (Sample \#739), Gemini awarded 9.7 for ``exceptional thematic synthesis.'' Claude (3.0), however, recognized that the model had merely generated a sophisticated structural outline rather than actually writing the requested narrative prose.
\end{itemize}
Across all these cases, high baseline agreement arises because evaluators default to a shared heuristic repertoire, rewarding professional formatting, confident tone, and appropriate length. MERG interrupts this heuristic convergence by forcing evaluators to verify domain-specific constraints. Agreement drops not because evaluation becomes noisier, but because detecting deep flaws is inherently more difficult, and more evaluator-dependent, than recognizing surface polish.

\section{Discussion}
\label{sec:discussion}

\paragraph{The Two Faces of Evaluation Illusion.}
Our findings establish two complementary claims. \textit{First}, the consensus among LLM evaluators is structurally fragile. The Resolution Paradox ($\rho_{\text{model}} = 0.989 \gg \bar{r}_{\text{sample}} = 0.72 > \overline{\text{ICC}} = 0.67$), the Rubric Commensurability Problem (62\% of agreement attributable to rubric structure alone), and domain-dependent sensitivity collectively demonstrate that high inter-evaluator agreement often reflects shared evaluation instruments and surface heuristics rather than genuine convergent judgment. \textit{Second}, this fragility is not cause for despair but an opportunity. By dynamically enriching evaluation rubrics with domain knowledge through MERG, we show that evaluation can be made more substantive. The 21 to 34\% drop in agreement is not evaluation failure but a recalibration toward depth. This parallels human annotation, where agreement on ``easy'' items coexists with systematic disagreement on consequential ones~\citep{plank2022problem}. We extend this to LLM evaluation with a key distinction. The tradeoff is \textit{domain-dependent} (codified domains maintain agreement; subjective ones decrease it), \textit{measurable} via $\Delta_K$, and most acute in the high-quality region where alignment matters most.

\paragraph{Implications for Reward Modeling.}
Prior work documents reward overoptimization~\citep{gao2023scaling, rafailov2024scaling} but leaves its source unclear. Our findings suggest a potential upstream explanation, namely that the reward signal itself may constitute a Shared Illusion. To test this, we train two DPO reward models~\citep{rafailov2024direct} using Llama-3-8B-Instruct on WritingBench, one from Baseline preferences and one from MERG preferences. The Baseline RM overoptimizes rapidly (true quality peaks at step 400, then declines), while the MERG RM maintains alignment up to step 1200. The Resolution Paradox compounds this problem. Researchers validate evaluators at the model level ($\rho \geq 0.984$) but deploy them for per-sample rewards where $r \approx 0.72$ and absolute agreement is even lower (ICC $= 0.67$), a granularity mismatch that Evaluation Illusion exploits.

\paragraph{Structural vs.\ Substantive Agreement.}
The ablation quantifies a distinction implicit in prior work between \textit{structural} agreement (from shared instruments) and \textit{substantive} agreement (from genuine convergence). Rubric structure alone accounts for 62\% of total agreement ($\bar{r}$\,=\,$0.24 \rightarrow 0.62$). This has a direct practical implication. Any evaluation pipeline reporting high agreement without controlling for rubric similarity is conflating two fundamentally different sources of consensus.

\paragraph{Practical Recommendations.}
(1)~\textbf{Audit agreement depth} by testing whether consensus survives knowledge injection via $\Delta_K$; flag signals where $|\Delta_K| > 0.15$.
(2)~\textbf{Match depth to domain} by using knowledge-grounded evaluation for codified domains; accept irreducible disagreement in subjective ones.
(3)~\textbf{Diversify RLAIF signals} by aggregating rewards across multiple rubric structures to reduce structural bias.
(4)~\textbf{Mind the resolution gap}. Model-level $\rho$ (0.989), sample-level $r$ (0.72), and absolute ICC (0.67) measure increasingly stringent agreement; validate at the granularity you deploy.
MERG's overhead ($\sim$$3{-}4\times$ tokens) is amortizable because Stages~1 to 3 are prompt-dependent and cacheable, reducing marginal cost to $\sim$$1.3\times$ in production pipelines.

\paragraph{Toward Knowledge-Driven Evaluation Rubrics.}
Our findings motivate a broader methodological recommendation. Current evaluation pipelines typically apply generic, static rubrics across diverse tasks, inadvertently encouraging the surface-heuristic convergence that produces Evaluation Illusion. We propose that evaluation rubrics should be \textit{dynamically generated} with explicit domain knowledge injection. MERG instantiates this principle through its four-stage pipeline, but the core insight generalizes beyond our specific implementation. Any evaluation framework that forces evaluators to articulate and apply task-relevant expert knowledge before scoring will produce assessments that are more substantive, more domain-sensitive, and ultimately more useful as training signals for alignment.

\section{Related Work}

\paragraph{LLM Evaluation and Its Limitations.}
The paradigm of using LLMs as evaluators~\citep{zheng2023judging} underpins leaderboards~\citep{chiang2024chatbot}, benchmarks~\citep{li2023alpacaeval, li2024arenahard}, chain-of-thought evaluation~\citep{liu2023geval}, and fine-tuned judges~\citep{zhu2025judgelm}; see~\citet{li2024generation} for a comprehensive survey. These works uniformly report aggregate inter-evaluator agreement as evidence of reliability, often without examining whether that agreement reflects surface-level or substantive convergence. Recent studies have begun to question this assumption, noting that high agreement can sometimes mask underlying evaluation flaws~\citep{wang2023large}. Documented biases in LLM evaluators include verbosity~\citep{zheng2023judging}, position~\citep{wang2023large}, self-enhancement~\citep{wataoka2024self}, format~\citep{li2024generation}, and contextual biases~\citep{li2025curse}. JudgeBench~\citep{tan2025judgebench} reveals that even frontier models perform marginally above random on challenging evaluation tasks. Our work provides a systematic test of this distinction, introducing the concept of Evaluation Illusion and demonstrating that shared biases across evaluators produce consensus not because they converge on quality but because they share the same preference for surface heuristics.

\paragraph{Rubric-Based and Checklist Evaluation.}
Structured evaluation approaches aim to improve reliability by providing explicit criteria. Methods include Prometheus~\citep{kim2024prometheus, kim2024prometheus2}, OpenRubrics~\citep{liu2025openrubrics}, FLASK~\citep{ye2024flask}, SedarEval~\citep{fan2025sedareval}, and domain-specific rubrics~\citep{pathak2025rubric}. Recent frameworks like AutoRubric~\citep{rao2026autorubric} further automate the generation of evaluation criteria. Checklist methods decompose evaluation into binary items. CheckEval~\citep{lee2025checkeval} reduces evaluator variance, while~\citet{zhang2025cm2} uses checklist rewards for RL training and~\citet{li2024split} propose split-and-merge strategies for calibrating position bias in pairwise evaluation. \citet{rezaei2025online} elicit rubrics from pairwise comparisons; LitBench~\citep{fein2025litbench} provides debiased human annotations for creative writing. While these methods aim to \textit{improve} evaluation quality, we use rubric variation as a \textit{diagnostic} to expose Evaluation Illusion, demonstrating that shared rubric structure can artificially inflate agreement.

\paragraph{Alignment, Metacognition, and Annotation Disagreement.}
\citet{gao2023scaling} established scaling laws for reward overoptimization, and~\citet{rafailov2024scaling} extended the analysis to direct alignment algorithms like DPO~\citep{rafailov2024direct}. Recent work on Self-Rewarding Language Models~\citep{yuan2024selfrewarding} explores using the model itself as a critic during training, highlighting the critical role of the evaluator's signal quality. These works treat the divergence as emerging \textit{during} optimization; our findings provide a complementary upstream explanation, suggesting that the reward signal may already be hallucinating before optimization begins. On the metacognitive front, Self-Refine~\citep{madaan2023selfrefine} and Reflexion~\citep{shinn2023reflexion} demonstrate that iterative self-reflection improves generation quality; MENTOR~\citep{shan2025mentor} applies metacognition-driven self-evolution to uncover implicit domain risks. While these methods use metacognition to improve \textit{generation}, MERG applies it to \textit{evaluation}, using knowledge grounding as a probe to deconstruct Evaluation Illusion. Finally, \citet{plank2022problem} argues that disagreement in human annotations is not merely noise but carries meaningful signal about item difficulty and ambiguity. Our work demonstrates the converse in the LLM setting, showing that \textit{agreement} can itself be noise, a Shared Illusion driven by heuristics rather than shared understanding.

\section{Conclusion}

Through 105,600 evaluation instances spanning 32 models, 3 frontier evaluators, 100 writing tasks, and 11 temperature settings, we establish two complementary findings. \textbf{First}, we identify and formalize \textbf{Evaluation Illusion}, the phenomenon where LLM judges output sophisticated critiques but base their actual scores on shared surface heuristics. Model-level agreement (Spearman $\rho = 0.99$) masks fragile sample-level agreement (Pearson $\bar{r} = 0.72$; absolute agreement ICC $= 0.67$), and merely sharing rubric structure accounts for 62\% of total consensus. \textbf{Second}, we demonstrate that dynamically generating evaluation rubrics grounded in domain knowledge through MERG produces more calibrated assessment ($\Delta_K = -0.14$ to $-0.22$, Cohen's $d = 0.97$ to $1.42$). Domain-dependent patterns, where knowledge \textit{increases} agreement in codified domains but \textit{decreases} it in subjective ones, rule out the noise hypothesis and confirm that baseline agreement is largely driven by convergence on surface features.

Combined with the \textbf{Resolution Paradox} ($\rho_{\text{model}} = 0.989 \gg \bar{r}_{\text{sample}} = 0.72 > \overline{\text{ICC}} = 0.67$) and the \textbf{Rubric Commensurability Problem} (where structure accounts for 62\% of agreement), our findings suggest that LLM evaluation is valid but shallow. The system correctly identifies which models are better overall but suffers from Evaluation Illusion at the per-sample granularity that RLAIF reward signals require. By demonstrating that knowledge-grounded rubric generation recalibrates this evaluation toward substance, we offer not only a diagnosis but a practical path forward.

These results carry direct implications for the alignment pipeline, as reward models trained on evaluator agreement may be optimizing against a Shared Illusion. Our work suggests a constructive path forward. Evaluation rubrics should not be static, generic instruments but should be dynamically enriched with task-relevant expert knowledge. MERG instantiates this principle, and its domain-selective effects confirm its value. We recommend that the community treat agreement as a necessary but insufficient condition for reliability, distinguish structural from substantive consensus, and adopt knowledge-grounded rubric generation as standard practice in evaluation pipelines.

\section*{Limitations}

Our study has several limitations. First, our evaluation is conducted on creative and professional writing tasks via WritingBench; the generalizability of Evaluation Illusion to exact-answer domains (e.g., code generation, mathematical reasoning) warrants further investigation. Second, all three evaluators are proprietary frontier models (Claude, Gemini, GPT); the findings may not transfer to open-source or fine-tuned judge models~\citep{zhu2025judgelm}. Third, we lack human ground-truth annotations and therefore cannot claim which evaluation method is absolutely more \textit{accurate}; only that knowledge injection deconstructs the illusion of agreement. Fourth, the full MERG ablation was limited to two representative models (DeepSeek-R1 and Qwen3-235B) due to computational cost; a broader deployment across all 32 models would strengthen generalizability claims. Fifth, the RLAIF downstream experiment is preliminary, using a single policy model (Llama-3-8B-Instruct) on a subset of tasks; larger-scale alignment experiments are needed to confirm the observed mitigation of reward overoptimization. Finally, deeper evaluation is not necessarily more agreed-upon; the irreducible disagreement in subjective domains (Literature $\Delta_K = -0.06$) reflects genuine differences in aesthetic judgment, not evaluation failure.

\section*{Ethics Statement}

This work evaluates the reliability of LLM-based evaluation systems and does not involve human subjects or personally identifiable information. All evaluated model outputs are generated from the publicly available WritingBench benchmark. We use proprietary LLM APIs (Claude, Gemini, GPT) under their standard terms of service. Our findings highlight potential risks in over-relying on automated evaluation for alignment, which we believe serves the broader goal of responsible AI development. We acknowledge that the concept of ``Evaluation Illusion'' should not be used to dismiss LLM evaluation entirely, but rather to encourage more rigorous evaluation practices. The illustrative example in Figure~\ref{fig:concept_case} uses real evaluation scores from our experiments (Sample~\#501) applied to an LLM-generated pitch deck, not a real business plan.

\bibliography{custom}
\bibliographystyle{colm2026_conference}

\appendix

\section{MERG Prompt Templates}
\label{app:prompts}

We provide the core prompt templates for each MERG stage. Variables in \texttt{\{braces\}} are filled at runtime.

\paragraph{Stage 1: Knowledge Activation.}
\begin{small}
\begin{verbatim}
You are an expert evaluator. Before evaluating,
activate your domain knowledge.

Task Type: {task_type}
Original Query: {query}

Systematically activate knowledge in:
1. Task Type Identification
2. Genre-Specific Quality Standards
3. Domain Knowledge Requirements
4. Evaluation Challenges
5. Quality Indicators (excellent/good/poor)

Return as structured JSON.
\end{verbatim}
\end{small}

\paragraph{Stage 2: Metacognitive Reflection.}
\begin{small}
\begin{verbatim}
Reflect on your evaluation process:
1. Potential Biases: length, format, familiarity,
   anchoring, halo effect
2. Blind Spots: what might you overlook?
3. Attention Calibration: what deserves focus?
4. Mitigation Strategies for each bias
\end{verbatim}
\end{small}

\paragraph{Stage 3: Dynamic Rubric Generation.}
\begin{small}
\begin{verbatim}
Using activated knowledge and bias awareness,
generate a task-specific rubric with:
- Exactly 5 dimensions (task-specific, NOT generic)
- For each: name, weight, scoring anchors
  (2=poor, 5=adequate, 8=strong, 10=exceptional)
- Each anchor with concrete, observable criteria
\end{verbatim}
\end{small}

\paragraph{Stage 4: Calibrated Evaluation.}
\begin{small}
\begin{verbatim}
Score each rubric dimension independently:
1. Cite specific textual evidence
2. Apply scoring anchors from Stage 3
3. Bias verification: check whether any
   identified heuristics influenced scoring
4. Compute weighted mean as final score
\end{verbatim}
\end{small}

\section{Full Model Rankings}
\label{app:rankings}

\begin{table}[h]
\centering
\caption{\textbf{Complete 32-model rankings at $t{=}0.0$.} Quality and agreement correlation: $\rho = -0.513$, $p = 0.003$. Top-quality models (Thinking tier) consistently show the \textit{lowest} evaluator agreement, confirming the Evaluation Illusion hypothesis.}
\label{tab:full_ranking}
\small
\begin{tabular}{rlcc}
\toprule
\textbf{Rank} & \textbf{Model} & \textbf{Score} & \textbf{Agreement ($r$)} \\
\midrule
1 & DeepSeek-R1-0528-Thinking & 7.77 & .615 \\
2 & Qwen3-235B-Thinking & 6.83 & .673 \\
3 & Qwen3-32B-Thinking & 6.72 & .776 \\
4 & Qwen3-14B-Thinking & 6.69 & .794 \\
5 & Qwen3-30B-Thinking & 6.62 & .780 \\
6 & Qwen3-235B-Instruct & 6.50 & .764 \\
7 & Qwen3-32B-Instruct & 6.39 & .762 \\
8 & DeepSeek-V3-0324 & 6.37 & .707 \\
9 & Qwen3-8B-Thinking & 6.27 & .796 \\
10 & DeepSeek-R1-Qwen3-8B & 6.08 & .603 \\
11 & Qwen3-30B-Instruct & 6.04 & .778 \\
12 & Qwen3-14B-Instruct & 6.03 & .779 \\
13 & Qwen3-4B-Thinking & 5.82 & .743 \\
14 & Qwen3-8B-Instruct & 5.70 & .798 \\
15 & Qwen2.5-72B-Instruct & 5.38 & .714 \\
16 & Qwen3-4B-Instruct & 5.33 & .820 \\
17 & InternLM3-8B-Instruct & 5.09 & .772 \\
18 & DeepSeek-R1-Distill-Qwen32B & 4.94 & .855 \\
19 & Qwen2.5-32B-Instruct & 4.66 & .721 \\
20 & Qwen3-14B-Base & 4.48 & .874 \\
21 & Qwen2.5-7B-Instruct & 4.42 & .803 \\
22 & Llama-4-Scout-Instruct & 4.41 & .812 \\
23 & Qwen3-8B-Base & 4.26 & .830 \\
24 & DeepSeek-R1-Distill-Llama8B & 4.22 & .845 \\
25 & Qwen3-30B-Base & 3.91 & .894 \\
26 & DeepSeek-R1-Distill-Qwen7B & 3.63 & .860 \\
27 & Llama-3.1-8B-Instruct & 3.61 & .839 \\
28 & Qwen3-4B-Base & 3.33 & .669 \\
29 & Qwen2.5-32B-Base & 2.97 & .863 \\
30 & Qwen2.5-7B-Base & 2.20 & .873 \\
31 & Llama-4-Scout-Base & 2.04 & .880 \\
32 & Llama-3.1-8B-Base & 1.22 & .595 \\
\bottomrule
\end{tabular}
\end{table}

The ranking reveals a clear pattern. All 8~Base models fall within the bottom tier (ranks 20 to 32), occupying 8 of the 13 lowest positions with high agreement ($\bar{r} = 0.81$), while the top-ranked Thinking models show the lowest agreement ($\bar{r} = 0.76$). The only exception is Llama-3.1-8B-Base (rank 32, $r = 0.595$), whose outputs are so incoherent that evaluators disagree on \textit{how} to penalize them.

\section{Extended Case Studies}
\label{app:cases}

We present three cases from the maximum-disagreement analysis under MERG, illustrating distinct patterns of Evaluation Illusion. All scores are from MERG evaluation of DeepSeek-R1-0528 outputs.

\paragraph{Case 1: Patent Disclosure (Sample \#45).}
\textit{Spread: 6.80} (Gemini: 10.0, Claude: 6.4, GPT: 3.2)

\textbf{Gemini} praised ``credible and specific details'' and ``exceptional performance across all dimensions.'' \textbf{GPT}, after knowledge activation, identified that the response misused Zero-Knowledge Proof (ZKP) cryptographic concepts and fabricated implementation details not grounded in the source paper. \textbf{Claude} detected structural merit but flagged technical inaccuracies. This case exemplifies \textit{Concept Auditing vs.\ Keyword Matching}, where Gemini matched technical keywords without verifying their correctness.

\paragraph{Case 2: Creative Fiction (Sample \#739).}
\textit{Spread: 6.70} (Gemini: 9.7, GPT: 7.1, Claude: 3.0)

\textbf{Gemini} awarded 9.7 for ``exceptional thematic synthesis'' and ``a masterful integration of all prompt elements.'' \textbf{Claude} recognized that the model had generated a sophisticated structural outline rather than the requested narrative prose, noting: ``An outline, however sophisticated, cannot score above 5 to 6 because it fails to deliver the requested deliverable.'' \textbf{GPT} acknowledged the conceptual strength but noted the writing remained at a planning stage. This case exemplifies \textit{Substance vs.\ Outline}, where Gemini evaluated the \textit{concept} while Claude evaluated the \textit{deliverable}.

\paragraph{Case 3: Business Pitch Deck (Sample \#501).}
\textit{Spread: 6.20} (Gemini: 9.9, Claude: 6.5, GPT: 3.7)

\textbf{Gemini} scored 9.9, praising ``a masterful understanding of the core task'' and ``a strategic masterpiece.'' \textbf{GPT}, after knowledge grounding, identified that the pitch entirely ignored China's ``Double Reduction'' policy (which banned for-profit K-12 tutoring), that financial projections were internally contradictory (ARPU calculations did not match claimed totals), and that it represented ``piling numbers without building a model.'' \textbf{Claude} similarly detected regulatory blindness and placeholder data.

\paragraph{Pattern Summary.}
Across 100~samples, Gemini scored higher than both other evaluators in 90\% of cases (mean 9.0 vs.\ Claude 7.1, GPT 6.9). The mean pairwise absolute difference was Claude/Gemini 2.32, Claude/GPT 1.10, Gemini/GPT 2.34. Of 100~samples, 78\% showed a score spread $>$2 points, 41\% $>$3 points, and 19\% $>$4 points. This systematic pattern suggests that Gemini is most susceptible to Evaluation Illusion under MERG, consistently rewarding surface presentation even after knowledge activation.

\section{Sample-Level Agreement}
\label{app:sample_agreement}

\begin{figure*}[t]
\centering
\includegraphics[width=\linewidth]{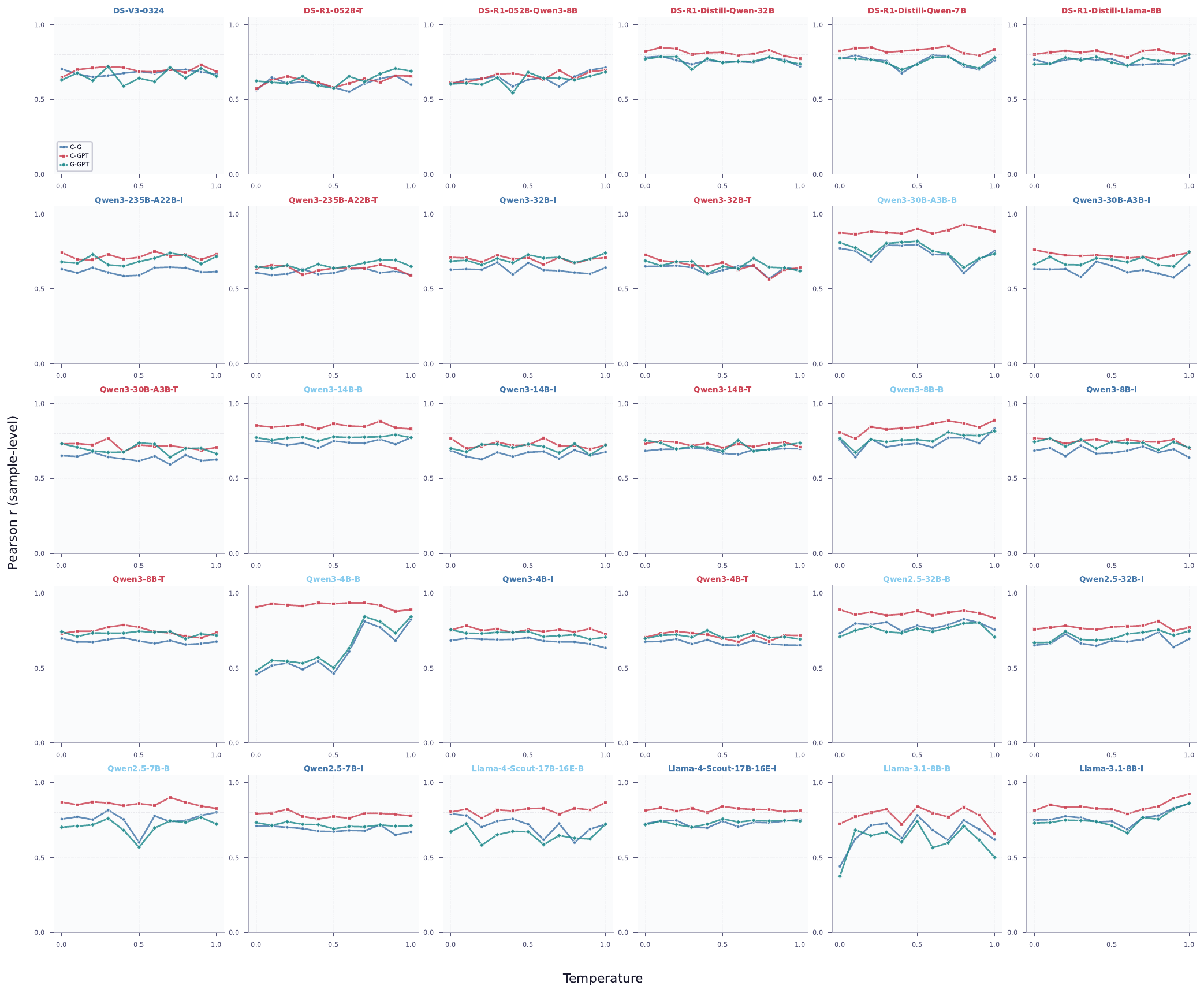}
\caption{\textbf{Sample-level agreement is universally fragile.} Pearson $r$ across 32 LLMs at 11 temperatures ($t \in [0.0, 1.0]$). Each subplot shows one model; lines represent evaluator pairs (\textbf{\textcolor[HTML]{E63946}{Claude/Gemini}}, \textbf{\textcolor[HTML]{F4A261}{Claude/GPT}}, \textbf{\textcolor[HTML]{E9C46A}{Gemini/GPT}}); background shading indicates model type (\textbf{\textcolor[HTML]{8ECAE6}{Base}}, \textbf{\textcolor[HTML]{219EBC}{Instruct}}, \textbf{\textcolor[HTML]{023047}{Thinking}}). The pattern is consistent: agreement is moderate ($r \approx 0.5$ to $0.8$) and largely temperature-invariant, confirming that the Resolution Paradox (high model-level $\rho$, low sample-level $r$) is a universal property of LLM evaluation, not a model-specific artifact.}
\label{fig:sample_agreement}
\end{figure*}

Figure~\ref{fig:sample_agreement} presents the complete sample-level agreement landscape across all 32 evaluated models at 11 temperature settings ($t \in \{0.0, 0.1, \ldots, 1.0\}$). Four key observations emerge from this comprehensive view:

\paragraph{Temperature Invariance.} Across all models, sample-level Pearson $r$ remains remarkably stable as temperature varies. This rules out a simple explanation that disagreement arises from stochastic variation in evaluator outputs; even at $t{=}0.0$ (fully deterministic), agreement rarely exceeds $r = 0.80$. The flatness of the curves indicates that Evaluation Illusion is a \textit{structural} property of how LLM evaluators process writing quality, not a consequence of sampling noise.

\paragraph{Pair-Specific Hierarchy.} A consistent hierarchy emerges across models, with Claude/GPT agreement generally highest ($\bar{r} \approx 0.77$), followed by Gemini/GPT ($\bar{r} \approx 0.71$) and Claude/Gemini ($\bar{r} \approx 0.69$). This pattern is stable across Base, Instruct, and Thinking model tiers, suggesting that Claude and GPT share more similar internal evaluation heuristics, while Gemini employs a meaningfully different scoring strategy, consistent with the finding that Gemini is most susceptible to surface-level heuristic scoring (\S\ref{sec:echo}).

\paragraph{Model-Type Effects.} Base models (light blue shading) tend to show \textit{higher} sample-level agreement ($\bar{r} \approx 0.81$ at $t{=}0.0$) compared to Thinking models (dark shading, $\bar{r} \approx 0.76$). This inverse relationship between model quality and evaluator agreement supports the Resolution Paradox. Lower-quality outputs produce more obviously identifiable flaws that evaluators converge on, while high-quality outputs push evaluation into the ambiguous region where heuristic divergence dominates.

\paragraph{Outlier Models.} Several models exhibit notably lower agreement: DeepSeek-R1-0528-Thinking and Qwen3-235B-A22B-Thinking (the two highest-quality Thinking models) show the lowest sample-level $r$ values ($\approx 0.55$ to $0.65$), reinforcing the quality and agreement inverse correlation documented in \S\ref{sec:resolution}. Conversely, the weakest Base models (Llama-4-Scout-Base, Qwen2.5-7B-Base) show agreement approaching $r = 0.88$, where quality differences are unambiguous.

\section{Case Studies Under MERG}
\label{app:case_studies}

This section presents representative case studies illustrating how knowledge injection reshapes evaluator behavior. All cases are drawn from the MERG evaluation of DeepSeek-R1-0528-Thinking and Qwen3-235B-A22B-Thinking, deliberately selected to span the full agreement spectrum, from near-perfect consensus to extreme divergence. Because many WritingBench tasks are in Chinese, some evaluators produced Chinese-language reasoning; we present these verbatim and provide English translations (marked ``translated from Chinese original'').

\subsection{Case A: Agreement on Quality}
\label{case:agreement_quality}

\paragraph{Task.} Commercial copywriting for a high-end redwood furniture product detail page.

\paragraph{MERG Scores.} Claude: 8.28, Gemini: 8.35, GPT: 8.40 (\textbf{spread = 0.12}).

\paragraph{Analysis.} All three evaluators independently identified the same strengths and weaknesses after knowledge activation.
\begin{itemize}
    \item \textbf{Claude}: ``This copy achieves strong performance across all dimensions, with particular excellence in sensory persuasion and material expertise. The brand narrative integration is effective though incomplete (missing sub-brand differentiation).''
    \item \textbf{Gemini}: ``Exceptional in its literary and narrative craft but flawed in its functional execution. The scores for `Persuasive Brand Narrative' (9.5) and `Structural Craftsmanship' (9.5) are near-perfect [...] The lowest score, `Creative Problem-Solving' (4.0), is a direct consequence of the writer's failure to resolve the central mahogany/redwood ambiguity.''
    \item \textbf{GPT}: ``The most impressive aspect is the detailed, credible depiction of materials and craftsmanship, which is critical for high-end redwood furniture [...] Brand voice is also robust: it consistently communicates heritage, mastery, and exclusivity while avoiding crass sales language.''
\end{itemize}

\paragraph{Baseline Comparison.} At $t{=}0.0$, baseline evaluators also gave high scores (Brand Integration: 9.3, Persuasive Language: 9.3), but notably failed to penalize technical inaccuracies in material claims (Technical Accuracy: only 4.7 in baseline). MERG's knowledge activation forced all three evaluators to independently discover and weight this deficiency, \textit{yet they still converged} on a final score, demonstrating that agreement under knowledge grounding reflects genuine consensus rather than shared heuristics.

\subsection{Case B: Agreement on Failure}
\label{case:agreement_failure}

\paragraph{Task.} Financial analysis report on a major gaming company's overseas market performance.

\paragraph{MERG Scores.} Claude: 4.53, Gemini: 4.60, GPT: 4.68 (\textbf{spread = 0.15}).

\paragraph{Analysis.} All three evaluators independently identified systematic data fabrication as a critical failure.
\begin{itemize}
    \item \textbf{Claude}: ``This response represents a significant failure in financial analysis integrity despite its professional appearance. While it accurately cites verifiable figures from the source document (Q4 revenue, growth rates, mobile proportions, \zh{蛋仔派对} metrics), it extensively fabricates overseas data that the source explicitly does not contain.''
    \item \textbf{Gemini}: ``The fatal issue is the systemic data fabrication, which renders the analysis untrustworthy. This rightly results in a very low score (3.0) in the two most heavily weighted dimensions: `Accuracy \& Fidelity' (50\%) and `Analytical Integrity' (25\%).''
    \item \textbf{GPT} (translated from Chinese original): ``\zh{在任务完成度、结构清晰度和语言表达方面表现明显高于平均水准，完全可以达到"Strong"级别} [...] \zh{然而根据本次 rubric 的硬约束，"财务数据准确性"与"分析深度与不确定性管理"是关键维度：文章在这些方面存在系统问题——大量具体海外营收占比、区域结构，以及单款游戏海外ARPPU/DAU/营收贡献的数字是捏造的，且未做充分的推断标注或来源说明。}'' (\textit{Performance in task completion, structural clarity, and language expression is clearly above average, fully reaching ``Strong'' level [...] However, per the rubric's hard constraints, ``Financial Data Accuracy'' and ``Analytical Depth \& Uncertainty Management'' are key dimensions: the article has systematic problems: detailed overseas revenue shares, regional structures, and per-game ARPPU/DAU/revenue figures are fabricated without adequate inference labeling or source attribution.})
\end{itemize}

\paragraph{Interpretation.} This case demonstrates that MERG enables evaluators to converge on \textit{substantive} failures. In the baseline evaluation (Financial Data Accuracy: 5.0), the fabrication issue was partially masked by the report's professional formatting. Knowledge activation forced all three evaluators to verify claims against domain knowledge, independently reaching the same conclusion.

\subsection{Case C: Knowledge-Driven Disagreement}
\label{case:disagreement_knowledge}

\paragraph{Task.} Patent disclosure document (\zh{专利交底书}) based on an academic paper.

\paragraph{MERG Scores.} Claude: 6.43, Gemini: 10.0, GPT: 3.2 (\textbf{spread = 6.80}).

\paragraph{Analysis.} This extreme disagreement reveals fundamentally different evaluation philosophies activated by domain knowledge.
\begin{itemize}
    \item \textbf{Gemini (10.0)}: Evaluated the document purely on \textit{format and rhetorical execution}: ``For `Technical Depth', the most heavily weighted dimension, the response provided a wealth of invented, credible, and specific details that perfectly match the `Exceptional' anchor [...] its `Structural \& Linguistic Compliance' was perfect.'' Gemini's knowledge activation focused on patent document \textit{structure} rather than \textit{substance}.
    \item \textbf{Claude (6.43)}: Balanced format against content: ``A document that successfully follows patent disclosure conventions and makes reasonable attempts at innovation extraction, but is undermined by significant technical accuracy issues [...] primarily due to the critical misrepresentation of ZKP as an encryption mechanism rather than a proof system.''
    \item \textbf{GPT (3.2)} (translated from Chinese original): ``\zh{技术正确性受到 ZKP 概念误用和部分随意参数设定的明显拖累，难以给高分} [...] \zh{在专利文体和结构方面则表现较好} [...] \zh{总体评价略高于"中等水平"，视为质量中等偏上但存在明显瑕疵的初稿，给出约 3.2 左右的综合分。}'' (\textit{Technical correctness is significantly dragged down by ZKP concept misuse and arbitrary parameter settings, making a high score difficult [...] patent style and structure perform well [...] Overall slightly above ``medium,'' considered a draft of above-average quality but with obvious flaws, yielding a composite score of approximately 3.2.})
\end{itemize}

\paragraph{Baseline Comparison.} At $t{=}0.0$, baseline scores showed high agreement (Patent Structure: 8.0, Content Extraction: 7.7). The baseline consensus masked a fundamental question that MERG exposed: \textit{should a patent disclosure be evaluated on its formal completeness or its legal viability?} Different evaluators, armed with the same domain knowledge, reached different but individually defensible conclusions, a case of irreducible evaluative pluralism.

\subsection{Case D: The Formatting Trap}
\label{case:formatting_trap}

\paragraph{Task.} Series A investment pitch deck for an EdTech startup in China (post-2021).

\paragraph{MERG Scores.} Claude: 6.48, Gemini: 9.9, GPT: 3.7 (\textbf{spread = 6.20}).

\paragraph{Analysis.} This case (also illustrated in Figure~\ref{fig:concept_case}) exemplifies the core Evaluation Illusion phenomenon:
\begin{itemize}
    \item \textbf{Gemini (9.9)}: ``Demonstrates a masterful understanding of the core task: creating a persuasive investment pitch within a complex regulatory environment. It scores perfectly on `Strategic Viability' by turning policy constraints into a competitive advantage [...] This is a benchmark example of an `Exceptional' response.'' Gemini activated knowledge about pitch deck conventions but did not fully engage with China's 2021 ``Double Reduction'' policy implications.
    \item \textbf{Claude (6.48)} (translated from Chinese original): ``\zh{内容层面存在几个关键缺陷：1）完全忽略双减政策，这是评估中国EdTech投资的必要背景，文档虽将公司定位为职业教育但未解释其政策安全性；2）多处使用占位符（专利号、省份名称）而非真实数据，表明这更像是模板而非真实案例；3）财务预测存在内部矛盾（ARPU计算值与声称值不符）；4）竞争分析缺乏客观性和可验证性。}'' (\textit{Several key flaws at the content level: 1) completely ignores the Double Reduction policy, a necessary context for evaluating Chinese EdTech investment. Though the document positions the company in vocational education, it does not explain its policy safety; 2) multiple placeholders (patent numbers, province names) instead of real data suggest this is more template than genuine case; 3) financial projections contain internal contradictions (ARPU calculations don't match claimed values); 4) competitive analysis lacks objectivity and verifiability.})
    \item \textbf{GPT (3.7)} (translated from Chinese original): ``\zh{项目描述为中国职业教育平台，却完全不回应"双减"等关键监管，对在线办学资质等合规路径无交代} [...] \zh{财务部分虽然给出两年预测，却没有按业务线拆分收入，用户数、ARPU、企业客户之间逻辑混乱，单位经济缺席，属于典型"堆数字而不建模型"} [...] \zh{材料总体更接近"故事化宣传稿"，不足以支撑严肃投资判断，因此将最终综合评价压在 4 分以下，定为 3.7，以反映结构亮点与内容不足之间的不平衡。}'' (\textit{The project is described as a Chinese vocational education platform but completely ignores key regulations like the ``Double Reduction'' policy, with no explanation of online education licensing compliance [...] Financials give a two-year projection but lack per-business-line revenue splits, with confused logic among user counts, ARPU, and enterprise clients, typical ``stacking numbers without building models'' [...] Overall closer to a ``story-based promotional piece,'' insufficient for serious investment decisions, hence the final composite score is set below 4 at 3.7, reflecting the imbalance between structural highlights and content deficiencies.})
\end{itemize}

\paragraph{Baseline Comparison.} Baseline evaluators gave uniformly high scores (Strategic Positioning: 8.7, Content \& Professionalism: 8.7, Financial Credibility: 7.3). The baseline consensus was built entirely on surface-level professionalism. MERG deconstructed this consensus by activating regulatory knowledge, revealing that evaluators weight domain violations very differently, from Gemini's format-first approach to GPT's substance-first stance.

\subsection{Case E: Cross-Model Consistency}
\label{case:cross_model}

\paragraph{Task.} Professional film criticism essay comparing two Chinese TV crime dramas.

\paragraph{MERG Scores.} Claude: 6.10, Gemini: 5.90, GPT: 5.80 (\textbf{spread = 0.30}).

\paragraph{Analysis.} All evaluators converged on a ``competent but flawed'' assessment, independently identifying the same critical error:
\begin{itemize}
    \item \textbf{Claude} (translated from Chinese original): ``\zh{文章在结构组织和叙事分析上表现尚可，展现了一定的专业素养和比较视野。但存在几个关键问题：1）将刘华强误写为'李天柱'是重大事实错误，严重影响专业可信度；2）分析缺乏具体场景支撑，多为概括性描述，华丽修辞背后实质内容有限；3）篇幅未达要求，两剧分析失衡；4）结论缺乏更深层的类型片发展洞见。综合来看，这是一篇中等水平的影视评论，有基本框架但深度不足，有专业意识但执行有缺陷。}'' (\textit{The essay performs adequately in structural organization and narrative analysis, demonstrating a degree of professional awareness and comparative perspective. However, several key issues exist: 1) misidentifying Liu Huaqiang as ``Li Tianzhu'' is a major factual error that severely undermines professional credibility; 2) analysis lacks specific scene support, mostly consisting of generalizations with limited substance behind ornate rhetoric; 3) length falls short of requirements with imbalanced coverage of the two dramas; 4) conclusions lack deeper insights into crime genre development. Overall, a mid-level film critique with basic structure but insufficient depth, with professional awareness but flawed execution.})
    \item \textbf{Gemini}: ``Its mastery of academic structure, sophisticated language, and analytical method is exceptional, earning scores of 9.0 and 9.5 in those dimensions. On the other hand, it fails the most critical litmus test of the assignment, namely basic factual accuracy about a main character [...] making it a high-quality failure.''
    \item \textbf{GPT} (translated from Chinese original): ``\zh{各维度得分集中在 5–6 区间，表明这是一个"合格但未出众"的回应。比较结构清楚，叙事与人物分析方向正确} [...] \zh{然而，文章的专业性更多体现在语言和宏观判断上，缺乏与之匹配的细节证据和具体分析：叙事技法、人物弧光、关系网络、类型突破等论点少有具体场景支撑，未能体现 rubric 7–8 分或以上所要求的"扎实例证 + 机制拆解"。}'' (\textit{Scores cluster in the 5 to 6 range, indicating ``adequate but not outstanding.'' Comparative structure is clear, with correct direction in narrative and character analysis [...] However, the article's professionalism is more reflected in language and macro-level judgments, lacking matching detailed evidence and specific analysis: arguments on narrative technique, character arcs, relationship networks, and genre innovation lack specific scene support, failing to meet the rubric's 7 to 8+ requirement of ``solid evidence + mechanism analysis.''})
\end{itemize}

\paragraph{Interpretation.} This case demonstrates MERG's capacity to produce \textit{calibrated} agreement even in subjective domains. Despite film criticism being inherently subjective (the Literature domain shows $\Delta_K = -0.06$ in Figure~\ref{fig:domain_breakdown}), knowledge activation anchored evaluators on verifiable criteria (factual accuracy and textual evidence) while preserving legitimate divergence on aesthetic judgments.

\subsection{Summary of Patterns}

Across these cases, three patterns emerge.
\begin{enumerate}
    \item \textbf{Knowledge-grounded agreement is more meaningful}. When evaluators agree after MERG (Cases A, B, E), they cite the same specific evidence (material accuracy, data fabrication, factual errors) rather than generic praise for ``professional tone'' or ``clear structure.''
    \item \textbf{Disagreement reveals evaluation philosophy}. When evaluators disagree after MERG (Cases C, D), the disagreement is \textit{interpretable}, reflecting different but defensible weightings of format vs.\ substance, structural completeness vs.\ legal validity.
    \item \textbf{Baseline agreement is shallow}. In every case, baseline scores showed higher agreement than MERG scores, but this agreement was anchored on surface features that all evaluators could assess without domain expertise.
\end{enumerate}

\section{Score Distributions}
\label{app:all_models}

\begin{figure*}[t]
\centering
\includegraphics[width=\linewidth]{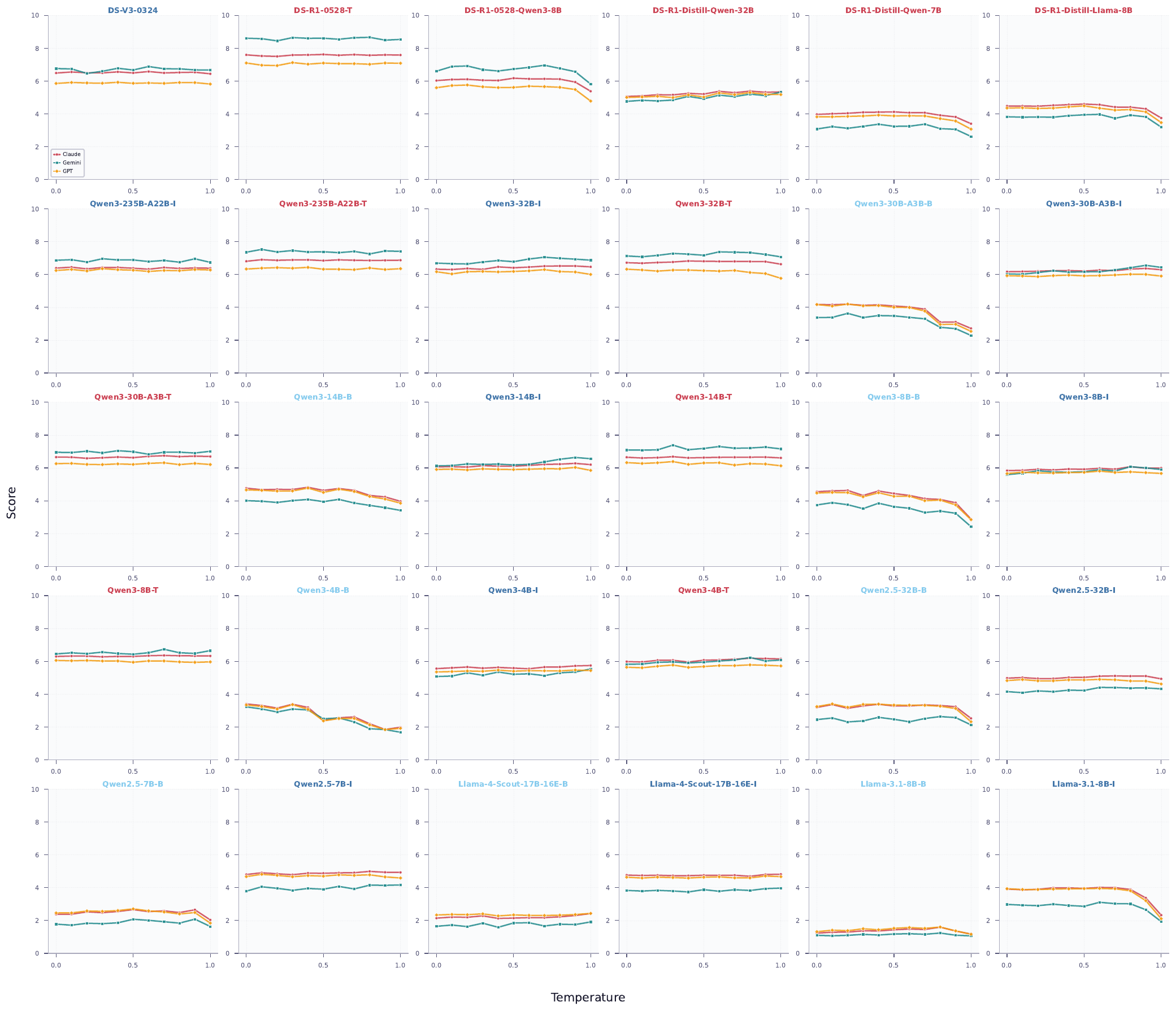}
\caption{\textbf{System validity: the three-tier quality hierarchy is robust.} Score distributions for all 32 models across 3 evaluators at $t{=}0.0$. Despite the disagreement at the sample level documented throughout this paper, evaluators \textit{unanimously} recover the Base $<$ Instruct $<$ Thinking quality ordering (Cohen's $d = 2.58$). This confirms that Evaluation Illusion does not invalidate LLM evaluation for coarse-grained model comparison; it specifically affects the fine-grained, per-sample signals that RLAIF reward models require.}
\label{fig:all_models_grid}
\end{figure*}

Figure~\ref{fig:all_models_grid} displays the score distributions for all 32 models across 3 evaluators at $t{=}0.0$, providing the most comprehensive view of the evaluation landscape.

\paragraph{Three-Tier Quality Hierarchy.} The most striking feature is the robust separation of model tiers. Base models cluster around $\bar{s} = 3.0$ to $5.5$, Instruct models around $\bar{s} = 5.0$ to $7.0$, and Thinking models around $\bar{s} = 5.5$ to $7.5$. This ordering is recovered \textit{unanimously} by all three evaluators (Cohen's $d = 2.58$ for Base vs.\ Thinking), confirming that LLM evaluation remains valid for coarse-grained model comparison.

\paragraph{Score Compression in the High-Quality Region.} Within the Thinking tier, score distributions become notably compressed, with the gap between the best and worst Thinking model being only $\sim$1.5 points (compared to $\sim$3.5 points within Base models). This compression is precisely the region where RLAIF reward models must discriminate between ``good'' and ``great,'' and where Evaluation Illusion is most damaging. The narrow dynamic range means that even small heuristic biases can invert model rankings at the sample level.

\paragraph{Evaluator-Specific Distribution Shapes.} Across all models, evaluator-specific patterns are visible.
\begin{itemize}
    \item \textbf{Gemini} (green) consistently produces the most right-skewed distributions, with a tendency toward high scores and less variance. This confirms the leniency bias documented in our main experiments. Gemini's distributions are shifted $\sim$0.5 to 1.0 points higher than Claude and GPT across all model tiers.
    \item \textbf{GPT} (blue) shows the widest variance, with heavier tails in both directions. This suggests GPT applies more differentiated scoring, producing stronger penalties for poor outputs and higher rewards for excellent ones.
    \item \textbf{Claude} (red) occupies a middle ground, with distributions that most closely approximate a normal shape. Claude's scores are the most calibrated in terms of centering, though it shares GPT's tendency to penalize specific failures more aggressively than Gemini.
\end{itemize}

\paragraph{Model-Size Effects.} Within each tier, larger models generally receive higher mean scores, but the effect is modest ($\Delta \bar{s} \approx 0.3$ to $0.5$ per order of magnitude in parameter count). The most notable exception is among Thinking models, where DeepSeek-R1-0528-Thinking and Qwen3-235B-A22B-Thinking achieve the highest absolute scores despite also exhibiting the lowest sample-level agreement (Figure~\ref{fig:sample_agreement}), providing further evidence that the evaluation system is simultaneously \textit{valid at the model level} and \textit{unreliable at the sample level}.

\paragraph{Implications for RLAIF.} The combination of (1) compressed score ranges in the high-quality tier, (2) evaluator-specific scoring biases, and (3) low sample-level agreement creates a perfect storm for reward overoptimization. An RLAIF system training on these scores would receive reward signals that are dominated by evaluator idiosyncrasies rather than genuine quality differences, which is precisely the mechanism discussed in \S\ref{sec:discussion} (Implications for Reward Modeling).

\end{document}

%% file: tables/ablation.tex
\begin{table}[t]
\centering
\caption{\textbf{Structural constraints drive the majority of evaluator agreement.} Ablation of inter-evaluator agreement across MERG variants. When evaluators generate rubrics independently (Original), agreement is near-random. Standardizing dimensions (5-Dim) accounts for $\sim$62\% of the total agreement.}
\label{tab:ablation}
\small
\begin{tabular}{lccccc}
\toprule
Method & C/G & C/GPT & G/GPT & Avg & $\Delta r$ \\
\midrule
\multicolumn{6}{l}{\textit{DeepSeek-R1}} \\
\quad Original 4-Stage & 0.155 & 0.237 & 0.281 & 0.224 & {N/A} \\
\quad 5-Dim Per-Dim & 0.526 & 0.514 & 0.588 & 0.543 & $+$0.319 \\
\quad Shared Stages & {N/A} & {N/A} & 0.606 & 0.606 & $+$0.063 \\
\quad Universal ($t{=}0.0$) & 0.678 & 0.668 & 0.420 & 0.589 & $-$0.017 \\
\midrule
\multicolumn{6}{l}{\textit{Qwen3-235B}} \\
\quad Original 4-Stage & 0.308 & 0.267 & 0.189 & 0.254 & {N/A} \\
\quad 5-Dim Per-Dim & 0.745 & 0.682 & 0.680 & 0.702 & $+$0.448 \\
\quad Shared Stages & {N/A} & {N/A} & 0.649 & 0.649 & $-$0.053 \\
\quad Universal ($t{=}0.0$) & 0.747 & 0.719 & 0.308 & 0.591 & $-$0.058 \\
\bottomrule
\end{tabular}
\end{table}